# Reasoning about Uncertainty in Metric Spaces

**Seunghwan Han Lee**
Mathematics Dept.
Indiana University Bloomington
Indiana, IN 47401
seunlee@cs.indiana.edu

## Abstract

We set up a model for reasoning about metric spaces with belief theoretic measures. The uncertainty in these spaces stems from both probability and metric structures. To represent both aspect of uncertainty, we choose an expected distance function as a measure of uncertainty. A formal logical system is constructed for the reasoning about expected distance. Soundness and completeness are shown for this logic. For reasoning on product metric spaces with uncertainty, a new metric is defined and shown to have good properties.

## 1 Introduction

For formal representation of uncertainty with probabilistic or belief theoretic measures, there exists a wide spectrum of models with complete logical systems (eg. see Halpern 2003, Nilsson 1986, Bacchus 1990, Gerla 1994). There also exists a wide variety of formal systems on spatial reasoning (eg. see Gabelaia *et al.* 2005, Asher & Vieu 1995, Lemon & Pratt 1998) and especially on reasoning about metric spaces (eg. see Kutz *et al.* 2003, Wolter & Zakharyaschev 2005). But it is hard to find a formal system when probabilistic uncertainty is present on metric spaces.

In probabilistic or statistical analysis, it is usually assumed that values of random variables are real numbers. Since most of statistical inference is related to reasoning about expectation and variance, restricting the range of random variables to the spaces where those values are well defined is considered as an acceptable sacrifice. A formal reasoning system on expectations with respect to probability measures and other belief theoretic measures can be found in (Halpern & Pucella 2002).

Even though expectations of an event cannot be defined on arbitrary metric spaces, the expectation of distances with respect to a fixed set is well defined in any metric space. For example, an expectation of a random location on a sphere cannot be defined, but an expected distance of a random location from the south hemisphere is a well defined notion. The expected distance is often interpreted as a loss function in statistical decision theory (Berger 1985).

$$Loss(a) = E_P(d(X, a)).$$

In this case, the distance function $d(x, a)$ with respected to a fixed point $a$ is used to calculate the cost for predicting $x$ when $a$ is a true state. A complimentary concept of a cost function in decision theory is a utility function. The expected utility function has been an important topic in economics for a long time (Von Neumann & Morgenstern 1944; Savage 1954; Schmeidler 1989). We can also find more generalized expected utility functions defined for plausibility measures and generalized utility functions. (Chu & Halpern 2003; 2004).

The objective of a decision problem is to find an action that minimizes a loss function. For example, if we have a uniform prior probability distribution on $[0, 1]$ where $d(x, y) = |x - y|$, the loss minimizing prediction would be $X = 0.5$. But in inference problems, the objective is to derive more information from given information. As an example, for the previous distance function, if we know that $E_P(d(X, 0)) = 1$, then we can deduce that $E_P(d(X, 1)) = 0$. For inference problems, a viewpoint from fuzzy logic is also meaningful. Let's assume that a fuzzy set $D_a$ represents a fuzzy concept "dissimilar from $a$" for a point $a$ in a 1-bounded metric space $\Omega$. A natural candidate for a membership function of $D_a$ would be $d(x, a)$. Given a probability distribution $P$, Zadeh (1968) introduces the probability of a fuzzy concept.

$$P(D_a) = \int_\Omega \mu_{D_a} dP = E_P(d(X, a)).$$

We can interpret this as an expected dissimilarity of $a$ from $\Omega - \{a\}$. Even though probability of fuzzy events is one possible representation of uncertainty in metric spaces, no formal reasoning system was provided yet. There has been various attempts to generalize Shannon's entropy to the metric spaces using expected similarity and expected distance (Yager 1992; Lee 2006).

In this paper, we will generate a formal axiomatic reasoning system using expected distance measures and prove soundness and completeness of the system. Since the expected distance function turns out to be a doubt function, we can also define various dual measures. For reasoning on product spaces, we will introduce a new product metric and define mutual independence with respect to expected distance.

## 2 Expected distance

A probability space is a tuple $(\Omega, \mathcal{F}, P)$ where $\Omega$ is a set, $\mathcal{F}$ is a $\sigma$-algebra of subsets of $\Omega$ and $P : \mathcal{F} \to [0, 1]$ is a function satisfying the properties of probability

- Prob1. $P(A) \geq 0$ for any $A \in \mathcal{F}$.
- Prob2. $P(\Omega) = 1$.
- Prob3. $P(\bigcap_{i=1}^{\infty} A_i) = \sum_{i=1}^{\infty} P(A_i)$ if $A_i \subset \mathcal{F}$ are disjoint.

To exclude problems with measurability, we will further assume that $\mathcal{F} = \mathcal{P}\Omega$. In most of applications $\Omega$ is a finite set and $P$ is defined at every point.

A 1-bounded pseudo metric space is a tuple $(\Omega, d)$ where $\Omega$ is a set and $d$ is a function on $\Omega \times \Omega$ that satisfies

- PMet1. $d(x, x) = 0$ (reflexive)
- PMet2. $d(x, y) = d(y, x)$ (symmetry)
- PMet3. $d(x, y) + d(y, z) \geq d(x, z)$ (triangular inequality)
- PMet4. $d(x, y) \leq 1$ (1-bounded)

for all $x, y, z \in \Omega$. To become a metric space, $d$ should satisfy an extra condition

$$\text{If } d(x, y) = 0 \text{ then } x = y.$$

In practice, this condition is too strong for a reasoning system. Allowing a zero distance between two instances is a common approach in spatial reasoning.

We further assume that the metric in our system is bounded by 1 because of following reasons. First, in most applications distance functions are practically bounded. Even if it is not bounded, there are conventional methods that normalize a metric space $(\Omega, d)$ into a 1 bounded metric space such as $(\Omega, \frac{d}{1+d})$. Especially if $d$ is bounded by $M$, $(\Omega, \frac{d}{M})$ is a straightforward conversion. Second, if metrics are not bounded by a certain number, it is hard to compare distances from multiple metrics. To get a normalized degree of uncertainty, a bound is a necessity. Third, 1 bounded metrics works well with probability or belief functions that are also 1 bounded. As an example, since $Pr(A)$ is bounded by 1, for independent events $A, B$, we have $Pr(A \text{ and } B) \leq Pr(A)$. Similar reasoning for expected distances when the distance functions are bounded.

**Definition 1.** *Let's assume that $(\Omega, \mathcal{F}, P, d)$ is a metric probability space with a probability measure $P$ on a $\sigma$-algebra $\mathcal{F} = \mathcal{P}\Omega$ and a 1 bounded pseudometric $d$. For a given subset $U \subset \Omega$, the expected distance of $U$ is defined as*

$$\begin{aligned} ed_{P,d}(U) &= \int_\Omega d(x, U) dP \\ &= \int_{\overline{U}} d(x, U) dP \end{aligned}$$

*where the distance between a point and a set is conventionally defined as*

$$d(x, U) = \min_{y \in U} d(x, y).$$

Since $d(x, \emptyset) = 1$ and $d(x, \Omega) = 0$ we have

$$ed_{P,d}(\emptyset) = 1 , \quad ed_{P,d}(\Omega) = 0 .$$

Even though an expected distance of a set $U$ is defined using $P$, when we perform an inference on $ed_{P,d}(U)$ we do not assume any knowledge of $P$. Therefore, we can safely omit $P$ from the notation in inference problems under unknown probability distribution $P$. We will also omit $d$ if there is no confusion for a base metric. Since we assumed that $\mathcal{F} = \mathcal{P}\Omega$, $d(x, U)$ is always a measurable function.

We can also define expected distance functions with respect to any kind of belief theoretic measures (Shafer 1976). The only difference is that the probability measure $P$ is changed into one of belief theoretic functions. A belief function $Bel$ is a function that satisfies

- B1. $Bel(A) \geq 0$ for any $A \in \mathcal{F}$.
- B2. $Bel(\Omega) = 1$.
- B3. $Bel(\bigcup_{i=1}^n A_i) \geq$
  $\sum_{r=1}^n (-1)^{r+1} \sum_{\{I' \subset \{1,\cdots,n\} | |I'|=r\}} Bel(\bigcap_{i \in I'} A_i)$.

A doubt function and a plausibility function are defined from *Bel*.

$$Doubt(A) = Bel(A^c)$$

$$Pl(A) = 1 - Bel(A^c).$$

But they can be also defined as set functions on $\Omega$ that satisfy the following properties.

$$Doubt(\emptyset) = 1, Doubt(\Omega) = 0$$

$$Doubt(\bigcap_{i=1}^{n} A_i) \geq \sum_{r=1}^{n} (-1)^{r+1} \sum_{\{I \subset \{1,2,\cdots,n\} | |I|=r\}} Doubt(\bigcup_{i \in I} A_i)$$

$$Plaus(\emptyset) = 0, Plaus(\Omega) = 1$$

$$Plaus(\bigcap_{i=1}^{n} A_i) \leq \sum_{r=1}^{n} (-1)^{r+1} \sum_{\{I \subset \{1,2,\cdots,n\} | |I|=r\}} Plaus(\bigcup_{i \in I} A_i)$$

## 3 Inclusion-exclusion principle

The inclusion-exclusion principle about set union and intersection is one of the most important properties in combinatorics, and has applications in diverse areas. Especially belief theoretic functions of Dempster-Shafer theory are defined by modifying the equality of the inclusion exclusion principle (Shafer 1976). We will propose a theorem on minimum and maximum that is similar to the inclusion-exclusion principle.

**Theorem 1 (Alternating min-max).** *Let $(G, \leq, *)$ be a commutative group with a linear order $\leq$ and a group operation $*$. For $d_1, \cdots, d_n \in G$ we have*

$$\max(d_1, \cdots, d_n) = \sum_{r=1}^{n} \Big[ \sum_{\{I \subset \{1,\cdots,n\} | |I|=r\}} \min_{i \in I} d_i \Big]^{(-1)^{r+1}}$$

$$\min(d_1, \cdots, d_n) = \sum_{r=1}^{n} \Big[ \sum_{\{I \subset \{1,\cdots,n\} | |I|=r\}} \max_{i \in I} d_i \Big]^{(-1)^{r+1}}$$

*where $-1$ represents the inverse in the group and the summation is for the group operation.*

The proofs can be found in (Lee 2006). Let's consider a special case when $G = \mathbb{R}$ and $* = +$ to get the following theorem.

**Theorem 2.** *Let $(\Omega, d)$ be a metric space. For a finite class $\{A_i\}_{i=1}^{n}$ of subsets of $\Omega$ we have*

$$d(x, \bigcap_{i=1}^{n} A_i) \geq \sum_{r=1}^{n} (-1)^{r+1} \sum_{\{I \subset \{1,\cdots,n\} | |I|=r\}} d(x, \bigcup_{i \in I} A_i)$$

*Proof.* From the definition of set distance, we have

$$d(x, \bigcap_{i \in I} A_i) \geq \max_{i \in I} d(x, A_i),$$

$$d(x, \bigcup_{i \in I} A_i) = \min_{i \in I} d(x, A_i).$$

The theorem follows from theorem 1. □

We have the following general fact of expectations for any belief theoretic measures since finite sums can be interchangeable with integrals.

**Theorem 3.** *Let $(\Omega, P)$, be a space with a belief theoretic measure $P$. If measurable functions $f$, $f_i$, $i \in I$ satisfy an (in)equality*

$$f \triangle \sum_{i \in I} a_i f_i$$

*where $\triangle$ is an (in)equality and $a_i$, $i \in I$ are constants, then*

$$E_P(f) \triangle \sum_{i \in I} a_i E_P(f_i).$$

We have the following property of the expected distance function as a special case of previous two theorems.

**Theorem 4.** *Let $P$ be a belief theoretic function. Given $A_i \subset (\Omega, P, d)$, $1 \leq i \leq n$, we have*

$$ed_{P,d}(\bigcap_{i=1}^{n} A_i)$$
$$\geq \sum_{r=1}^{n} (-1)^{r+1} \sum_{\{I \subset \{1,2,\cdots,n\} | |I|=r\}} ed_{P,d}(\bigcup_{i \in I} A_i)$$

This theorem shows that expected distance functions become doubt functions. An important property of belief theoretic functions is the Möbius inversion (Shafer 1976). A mass function $m$ is defined to be a set function on $\Omega$ that satisfies

1. $m(\emptyset) = 0$
2. $\sum_{U \subset \Omega} m(U) = 1$.

It is known that the following inversion theorem holds for any doubt function.

**Theorem 5.** *If ed is an expected distance function, there is a mass function m such that*

$$ed(A) = \sum_{U \subset A^c} m(U)$$

$$m(A) = \sum_{U \subset A} (-1)^{|A|-|U|} ed(\overline{U})$$

At this point we note that our inequalities are fundamentally different from those in Halpern and Pucella's paper (2002) that can be stated as

$$E_{Bel}(\bigvee_{i=1}^n X_i)$$
$$\geq \sum_{r=1}^n (-1)^{r+1} \sum_{\{I \subset \{1,2,\cdots,n\} || I|=r\}} E_{Bel}(\bigwedge_{i \in I} X_i)$$

where $X_i$'s are gambles (random variables). Our inequality is derived from the inequality of distance functions. It does not change the direction regardless of base measures. The above inequality came from the belief theoretic measure upon which the expectation is defined. Therefore, different inequalities hold for different base measures. For example, the above inequality becomes an equality when the base measure is a probability measure.

## 4 Dual measures of the expected distance function

Since expected distance function is a doubt function, we can also consider dual functions of an expected distance function.

**Definition 2.** *The expected similarity, expected absoluteness, and expected relativeness are defined as*

$$\begin{aligned} es(A) &= 1 - ed(A) \\ ea(A) &= ed(A^c) \\ er(A) &= 1 - ed(A^C) \end{aligned}$$

Because $ed(A) + ed(A^c) \leq 1$ we have

$$\begin{aligned} ed(A) &\leq er(A) \\ ea(A) &\leq es(A) \end{aligned}$$

The equality holds when the distance function is crisp (values are 0 or 1) or when the probability measure is crisp. Let's see $ea$ in its integration form.

$$ea(A) = \int_A d(x, A^c) dP$$

This measures the expected distance between $A$ and $A^c$. The bigger $ea$ is, the more $A$ becomes an absolute choice. Therefore the bigger $er$ is, the less $A$ becomes an absolute choice. We can compare the meaning of $ed$ and $er$ in the following example.

**Example 1.** *Let's assume that a search team for a lost child needs to pick up one of two places A, B as a starting point. If $d(A, B)$ is small, it is not meaningful to spend time to optimize the starting point since the other point will be reached soon. This can be represented with expected distance and expected relativeness. Let $d(A, B) = 0.2$, and the real probability of the lost child being in A and B be $P(A) = 0.1$ and $P(B) = 0.9$. Then,*

$$ed(\{A\}) = 0.2 \cdot 0.9 = 0.18 .$$

*Because the cost of wrong prediction is low, even a relatively improbable place "A" has low expected distance. But the expected relativeness is big because the alternative "B" also has low expected distance.*

$$\begin{aligned} er(\{A\}) &= 1 - ed(\{A\}^c) \\ &= 1 - 0.2 \cdot 0.1 = 0.98 . \end{aligned}$$

*This represents that even though predicting "A" has little risk, it need not be the best choice since predicting $\{A\}^c = \{B\}$ also has little risk.*

As in belief theory we can consider a tuple

$$[ed(A), er(A)]$$

to represent the uncertainty of a probabilistic event $A$ in metric space. The interval becomes smaller when the distance function is crisp with respect to $A, A^c$. It means that the set $A$ becomes a more distinctive category with respect to the distance. If the interval is very small, an expected distance behaves like a probability measure.

$$ed(A) + ed(A^c) \approx 1 .$$

Since $ed(A) \leq P(A^c)$, actual values can be derived from probabilities.

$$ed(A) \approx P(A^c).$$

## 5 Inference on product of metric spaces

When we deal with inference problems with a large number of variables, it is almost impossible to maintain the data set without factoring the whole space into product spaces. To reduce the dimension of product spaces, statistical analysis often assumes that variables are independent. Events $A_1, \cdots, A_n$ are mutually independent relative to probability if and only if

$$Pr(\bigcap_{i \in I} A_i) = \prod_{i \in I} Pr(A_i)$$

for any $I \subset \{1, \cdots, n\}$. A set of $\sigma$-fields $\mathcal{F}_1, \cdots, \mathcal{F}_n$ are said to be mutually independent when for any given $A_i \in \mathcal{F}_i$, $A_1, \cdots, A_n$ are mutually independent. We can extend this condition to expected distance functions.

**Definition 3.** *Events $A_1, \cdots, A_n$ are mutually independent relative to expected distance if and only if*

$$1 - ed(\bigcap_{i \in I} A_i) = \prod_{i \in I}(1 - ed(A_i))$$

*for any $I \subset \{1, \cdots, n\}$.*

In probability theory, the product measure space and the product probability are defined as

$$\begin{aligned}
\prod_{i=1}^{n} \mathcal{F}_i &= \sigma\text{-field generated by} \\
&\quad \{\prod_{i=1}^{n} A_i | A_i \in \mathcal{F}_i\}, \\
(P_1 \times P_2)(E) &= \int_X P_2(E_x) \, dP_1(x) \\
&= \int_Y P_1(E_y) \, dP_2(y) .
\end{aligned}$$

where $E_x = \{y : (x,y) \in E\}$ and $E_y = \{x : (x,y) \in E\}$. This product probability makes $E_x, E_y$ independent events. We will construct a product expected distance function so that it also preserves the mutual independence. For that purpose we need a special product metric.

**Definition 4.** *If $\{(\Omega_i, d_i)\}_{1 \leq i \leq n}$ are 1 bounded metric spaces, a metric $\Lambda^n \vec{d}$ on $\prod_{i=1}^{n} \Omega_i$ is defined as*

$$\begin{aligned}
&\Lambda^n \vec{d}(\vec{x}, \vec{y}) \\
&= \sum_{r=1}^{n} (-1)^{r+1} \prod_{\{I \subset \{1,2,\cdots,n\} | |I|=r\}, i \in I} d_i(x_i, y_i) \\
&= 1 - \prod_{i=1}^{n}(1 - d_i(x_i, y_i))
\end{aligned}$$

*where $\vec{x}_n = (x_1, \cdots, x_n)$.*

**Theorem 6.** *If $\{(\Omega_i, d_i)\}_{1 \leq i \leq n}$ are 1 bounded metric spaces, then $(\prod_{i=1}^{n} \Omega_i, \Lambda^n \vec{d})$ becomes a 1 bounded metric space.*

Because of restricted space, refer to (Lee 2006) for omitted proofs. Most of known metrics on product spaces such as Euclidean metric, product metric, supremum metric, etc. are not appropriate for reasoning. The Euclidean metric does not even become a 1-bounded metric on a product space. Other metrics are dependent on the order of spaces or not constructed recursively. $\Lambda^n \vec{d}$ satisfies all such requirements. Furthermore, it is a unique distance function that satisfies following conditions.

**Theorem 7.** *$\Lambda^n \vec{d}$ is unique under the following conditions.*

1. $\Lambda^{n+1}(d_1, \cdots, d_n, d_{n+1})$
   $= \Lambda^2(\Lambda^n(d_1, \cdots, d_n), d_{n+1})$
2. $\Lambda^n(d_1, \cdots, 1, \cdots, d_n) = 1$
3. $\Lambda^n(0, \cdots, 0) = 0$
4. $\Lambda^n(d_1, \cdots, d_i, \cdots, d_j, \cdots d_n)$
   $= \Lambda^n(d_1, \cdots, d_j, \cdots, d_i, \cdots, d_n)$
5. $\frac{\partial \Lambda^n}{\partial d_i}(d_1, \cdots, d_n) = f(d_1, \cdots, \hat{d}_i \cdots, d_n)$

The following theorem shows that if more different aspects are known, the distance between two concepts becomes bigger.

**Theorem 8.** *Given $(\Omega, d_i)$, $1 \leq i \leq n+1$,*

$$\Lambda^n(d_1, \cdots d_n) \leq \Lambda^{n+1}(d_1, \cdots, d_n, d_{n+1})$$

A partial order $<$ on $[0, 1]^n$ can be defined as $\vec{x} < \vec{y}$ if and only if $x_i < y_i$ for some $i$, and not $x_i > y_i$ for any $0 \leq i \leq n$. $\Lambda^n \vec{d}$ preserves this order in the following sense.

**Theorem 9.** *If $\vec{x} < \vec{y} < \vec{z}$, then $\Lambda^n \vec{d}(\vec{x}, \vec{y}) < \Lambda^n \vec{d}(\vec{x}, \vec{z})$.*

Metrics such as $\sup(d_1, \cdots, d_n)$ does not preserve this order. Now we will prove the mutual independence under $\Lambda^n \vec{d}$.

**Lemma 10.** *For $A_i \subset (\Omega_i, d_i)$ we have*

$$1 - \Lambda^n \vec{d}(\vec{x}, \prod_{i=1}^{n} A_i) = \prod_{i=1}^{n}(1 - d_i(x_i, A_i)).$$

The following theorem for mutual independence follows from this lemma.

**Theorem 11.** *Let $A_i \subset (\Omega_i, \mathcal{F}_i, P_i, d_i)$ for each $i$. If $\mathcal{F}_1, \cdots, \mathcal{F}_n$ are mutually independent, then on $(\prod_{i=1}^{n} \Omega_i, \prod_{i=1}^{n} \mathcal{F}_i, P, \Lambda^n \vec{d})$*

$$1 - ed_{P, \Lambda^n \vec{d}}(\prod_{i \in I} A_i) = \prod_{i \in I}(1 - ed_{P_i, d_i}(A_i))$$

Given $A_i \subset \Omega_i$, it is a common convention to denote $A_i \subset \prod_{i=1}^{n} \Omega_i$ for $\Omega_1 \times \cdots \times \Omega_{i-1} \times A_i \times \Omega_{i+1} \times \cdots \times \Omega_n$.

**Lemma 12.** *If $A_i, B_i \subset (\Omega_i, d_i)$ for $i = 1, \cdots, n$ then*

$$\begin{aligned}
&\Lambda^n \vec{d}(A_1 \times \cdots \times \Omega_i \times \cdots \times A_n, \\
&\quad B_1 \times \cdots \times B_i \times \cdots \times B_n) \\
&= \Lambda^{n-1} \vec{d}(A_1 \times \cdots \times \hat{A}_i \times \cdots \times A_n, \\
&\quad B_1 \times \cdots \times \hat{B}_i \times \cdots \times B_n).
\end{aligned}$$

Because of this lemma, for $A_i, B_i \subset (\Omega_i, d_i)$ we have

$$d_i(A_i, B_i) = \Lambda^n \vec{d}(A_i, B_i).$$

**Theorem 13.** *Let $(\Omega_i, \mathcal{F}_i, P_i, d_i)$ be the probability space with metric. If $A_i \subset \Omega_i$ then*

$$ed_{P, \Lambda^n \vec{d}}(\Omega_1 \times \cdots \times A_i \times \cdots \times \Omega_n) = ed_{P_i, d_i}(A_i).$$

# 6 Axiomatizing expected distance

## 6.1 Syntax and semantics

The syntax of a logic of expected distance $\mathcal{L}^{ED}$ is defined as follows.

**Definition 5.** *Let $\Pi = \{P_1, P_2, \cdots\}$ be a set of* primitive propositions. *The set of* propositional formulas *is the closure of $\Pi$ under the Boolean operations $\wedge, \neg$ as in the propositional logic. An* expected distance term *is an expression of the form $a_1 ED(\varphi_1) + \cdots + a_n ED(\varphi_n)$, where $\varphi_i$ is a propositional formula and $a_i$'s are real numbers. A* basic expected distance formula *is a statement of the form $t \geq \alpha$ where $t$ is a expected distance term and $\alpha$ is a real number. An* expected distance formula *or just a* formula *of the language $\mathcal{L}_E$ is a Boolean combination of basic expected distance formulas.*

As an example, $(2ED(P \wedge Q) + 0.23ED(\neg Q) \geq 0.2) \vee (ED(\neg P) < 0.1)$ is a formula. We will use abbreviations such as $t < \alpha$ for $\neg(t \geq \alpha)$, $t \leq \alpha$ for $(-1)t \geq -\alpha$, $t = \alpha$ for $(t \leq \alpha) \wedge (t \geq \alpha)$.

The semantics of $\mathcal{L}^{ED}$ is defined on a probability structure with a metric, which is a tuple $M = (\Omega, \mathcal{F}, P, d, \pi)$. $(\Omega, \mathcal{F}, P)$ is a probability space where all subsets of $\Omega$ are measurable. $\pi : \Pi \to \{0, 1\}$ is a truth assignment function of atomic propositions. $d$ is a metric on $\Omega$. We first define the interpretation of a propositional formula $\phi$.

$$\varphi^M = \{\omega \in \Omega | \pi(\omega)(\varphi) = 1\}.$$

For a given model $M$,

$$M \models a_1 ED(\varphi_1) + \cdots + a_n ED(\varphi_n) \geq \alpha$$
$$\text{iff} \quad a_1 ed(\varphi_1^M) + \cdots + a_n ed(\varphi_n^M) \geq \alpha$$

We extend $\models$ to arbitrary formulas as

$$\begin{aligned} M \models f_1 \wedge f_2 & \quad \text{iff} \quad M \models f_1 \text{ and } M \models f_2 \\ M \models \neg f & \quad \text{iff} \quad M \not\models f \end{aligned}$$

Note that even though we defined a semantics for probability structures, since an expected distance can be taken with respect to any other belief theoretic measures, we can make logics for those structures just by changing the probability structure into those.

## 6.2 Axioms of $\mathcal{L}^{ED}$

Now we will construct a sound and complete axioms for our logic. First, we use the axioms $Axio_L$ that is defined in (Fagin, Halpern, & Megiddo 1990), (Fagin & Halpern 1994), and shown to be sound and complete.

I. $Axiom_P$: Axioms for propositional logic.
II. $Axiom_L$ : Axioms for linear inequalities including $Axiom_P$.

The following axioms characterize the logic of expected distances.

III. $Axiom_{ED}$: Axioms for expected distance including $Axiom_L$.

1. (true) $ED(true) = 0$.

2. (false) $ED(false) = 1$.

3. (nonnegative) $ED(\varphi) \geq 0$.

4. (inclusion-exclusion) $ED(\bigwedge_{i=1}^n \varphi_i)$
   $\geq \sum_{r=1}^n (-1)^{r+1} \sum_{\{I \subset \{1,2,\cdots,n\} | |I|=r\}} ED(\bigvee_{i \in I} \varphi_i)$.

5. (substitution) $\varphi \Leftrightarrow \psi$ in propositional logic, then $ED(\varphi) = ED(\psi)$.

The inclusion exclusion axiom is the generalization of the simple case

$$ED(\varphi \wedge \psi) \geq ED(\varphi) + ED(\psi) - ED(\varphi \vee \psi).$$

Even though these axioms are for probability measures, the axioms for other belief theoretic measures would be the same. We can prove the following properties of expected distance from $Axiom_{ED}$.

**Theorem 14.** *1. $ED(\neg \varphi) \leq 1 - ED(\varphi)$.*

*2. $ED(\varphi) \geq ED(\varphi \wedge \psi) + ED(\varphi \wedge \neg \psi) - 1$.*

*3. If $\varphi \Rightarrow \psi$ then $ED(\varphi) \leq ED(\psi)$.*

## 6.3 Soundness and Completeness

Now we will prove the soundness and completeness theorems for the logic of expected distance. Even though an expected distance function is a doubt function, the axioms for doubt function need not be complete for the system of expected distance. For example, a probability measure is a plausibility measure, but the axiom set of plausibility measure is not complete for probabilistic systems.

**Theorem 15.** *$Axiom_{ED}$ is sound and weakly complete.*

*Proof.* As for soundness, all axioms except for inclusion-exclusion axiom can be easily proved to be sound. The inclusion-exclusion axiom comes from theorem 4.

To prove the weak completeness we will build a model for any given consistent formula $f \in \mathcal{L}^{ED}$. Once we have the model existence, the completeness is proved as following. Let's assume that $\Gamma \models \varphi$ for a finite $\Gamma$. If $\Gamma \not\vdash \phi$ then $\bigwedge \Gamma \vee \neg \phi$ is consistent. By model existence we have a model $M$ such that $M \models \bigwedge \Gamma \vee \neg \varphi$. Therefore $\Gamma \not\models \varphi$. This contradiction solves the completeness.

Let's represent $f$ in a disjunctive normal form $g_1 \vee \cdots \vee g_n$ where each $g_i$ is a conjunction of basic expected distance formulas and their negations. If $f$ is consistent, then some $g_i$ is consistent. Moreover, any model that satisfies $g_i$ also satisfies $f$. Therefore we only need to make a model for the formula $g_i$.

Let $\{P_1, \cdots, P_k\}$ be the set of all primitive propositions that appears in $g_i$. Let $\{\varphi_1, \cdots, \varphi_{2^k}\}$ be the set of all possible formulas where $\varphi_i = \pm P_1 \wedge \cdots \wedge \pm P_k$. ($\pm P_j$ represents $P_j$ or $\neg P_j$). Let $n = 2^k$. For any propositional formula $\psi$ that appear in $g_i$, we have a set $I \subset \{1, \cdots, n\}$ such that

$$\models \psi \Leftrightarrow \bigvee_{i \in I} \varphi_i .$$

Let $g'_i$ be the formula made from $g_i$ by substituting all occurrences of propositional formula $\psi$ with equivalent $\bigvee_{i \in I} \varphi_i$'s. Because of the soundness of the axioms, it is enough to construct a model for the formula $g'_i$.

Now consider the formula $h$ that is the conjunction of $g'_i$ with the following formulas that represents the axioms. The idea is to add enough restrictions so that any solution of the linear equation satisfy conditions of expected distance.

We first conjunct a formula for true. Since $true \equiv \bigvee_{i=1}^n \varphi_i$,

$$ED(\bigvee_{i=1}^n \varphi_i) = 0 .$$

Second, we conjunct a formula for the empty set. Since $false \equiv \bigvee_{\emptyset} \varphi_i$,

$$ED(\bigvee_{\emptyset} \varphi_i) = 1 .$$

Third, we conjunct $2^n$ formulas that represent the nonnegative conditions for all $I \subset \{1, \cdots, n\}$.

$$ED(\bigvee_{i \in I} \varphi_i) \geq 0 .$$

Finally, we conjunct formulas that represent the inclusion-exclusion conditions. For all $K \subset \mathcal{P}\{0, \cdots, n\}$,

$$ED([\bigvee_{i \in \cap K} \varphi_i])$$
$$\geq \sum_{r=1}^{|K|} (-1)^{r+1} \sum_{\{K' \subset K | |K'|=r\}} ED(\bigvee_{i \in \cup K'} \varphi_i) .$$

Since $g'_i$ is consistent in $Axiom_{ED}$, $h$ is also consistent in $Axiom_{ED}$. Therefore it should be consistent in $Axiom_L$. In $Axiom_L$, the expected distance terms $ED(\bigvee_{i \in I} \varphi_i)$'s are considered as variables. Since we know the completeness of the $Axiom_L$, $h$ should have a model. A formula in the logic of linear inequalities is a system of linear inequalities. A model of a formula has an interpretation map that assigns a solution for the linear inequality system of the given formula. Therefore, the model of $h$ assigns a number $e_I$ for each variable $ED(\bigvee_{i \in I} \varphi_i)\}$. So that if the $e_I$'s are substituted into the variables, the linear inequality system is satisfied. So we have

$$\begin{aligned} e_\emptyset &= 1, \\ e_{\{1,\cdots,n\}} &= 0, \\ e_I &\geq 0, \\ e_{\cap K} &\geq \sum_{r=1}^{|K|} (-1)^{r+1} \sum_{\{K' \subset K | |K'|=r\}} e_{\cup K'} . \end{aligned}$$

From Lemma 16 there is a model $M = (\Omega, \mathcal{F}, P, d, \pi)$ such that

$$ed_{P,d}([\bigvee_{i \in I} \varphi_i]^M) = ed_{P,d}(\bigcup_{i \in I} \varphi_i^M) = e_I .$$

Therefore, $M \models h$ and the theorem is proved. □

**Lemma 16.** *For $\{e_I \in \mathbb{R} | I \subset \{1, \cdots, n\}\}$ such that*

$$\begin{aligned} e_\emptyset &= 1, \\ e_{\{1,\cdots,n\}} &= 0, \\ e_I &\geq 0, \\ e_{\cap K} &\geq \sum_{r=1}^{|K|} (-1)^{r+1} \sum_{\{K' \subset K | |K'|=r\}} e_{\cup K'} . \end{aligned}$$

*where $K \subset \mathcal{P}\{1, \cdots, n\}$, we have a probability metric space $M = (\Omega, \mathcal{F}, P, d, \pi)$ such that*

$$ed_{P,d}([\bigvee_{i \in I} \varphi_i]^M) = ed_{P,d}(\bigcup_{i \in I} \varphi_i^M) = e_I .$$

*Proof.* An underlying space $\Omega$ and a propositional interpretation $\pi$ is constructed as follows.

$$\begin{aligned}
\Omega &= \bigcup_i \varphi_i^M \\
\varphi_i^M = \pi(\varphi_i) &= X_i \cup Y_i \\
X_i &= \{x_{i,J}\}_{J \subset \{1,\cdots \hat{i} \cdots, n\}} \\
Y_i &= \bigcup_{j \neq i} \{y_{i,j,K}\}_{K \subset \{1,\cdots,\hat{j},\cdots,n\}} .
\end{aligned}$$

Before constructing a probability measure $P$, let's consider a discrete space of n-points, $\Omega' = \{a_{\varphi_1^M}, \cdots, a_{\varphi_n^M}\}$. A set function $ed' : \mathcal{P}\Omega' \to \mathbb{R}$ is defined as $ed'(\bigcup_{i \in I}\{a_{\varphi_i^M}\}) = e_I$. Because of the properties of $e_I$, $ee'$ satisfies all properties of doubt function. Therefore we can apply Möbius transformation to get the mass function $m : \mathcal{P}\Omega' \to \mathbb{R}$. If we represent $m(\bigcup_{i \in I}\varphi_i^M) = m_I$, we have the following equality from the definition of Möbius transformation.

$$m_I = \sum_{J \subset I} (-1)^{|I|-|J|} e_J .$$

$$e_I = \sum_{J \subset I^c} m_J .$$

Consider the measurable space $\mathcal{F} = \mathcal{P}\Omega$. Since $\Omega$ is finite, a probability distribution on $\mathcal{F}$ is determined by the probability at each point of $\Omega$. So, we can construct the probability distribution $P$ on $\mathcal{F}$ as

$$P(x_{i,J}) = \frac{m_{J^c}}{|J^c|}, \quad P(y_{i,J,K}) = 0 .$$

Now we will define a pseudo distance function $d$ on $\Omega$ as

$$d(x_{j',K'}, y_{i,j,K}) = \begin{cases} 0 & \text{if } j = j' \text{ and } K = K' \text{ and } i \notin K \\ 1 & \text{else} \end{cases}$$

$$d(y_{i,j,K}, y_{i',j',K'}) = \max(d(x_{j,K}, y_{i,j,K}), d(x_{j',K'}, y_{i',j',K'}))$$

$$d(x_{i,J}, x_{i',J'}) = 1$$

Let's show the triangular inequality. Since all values are 0 or 1, the only cases that we are concerned is the case of $0 + 0 \geq 1$. Since 0 distance can not happen between $x_{i,J}$'s, the remaining cases are as follows. First,

$$d(x_{j,K}, y_{i,j,K}) + d(x_{j,K}, y_{i,j',K}) \geq d(y_{i,j,K}, y_{i',j,K}) ,$$

$$d(x_{j,K}, y_{i,j,K}) + d(y_{i,j,K}, y_{i',j,K}) \geq d(x_{j,K}, y_{i',j,K})$$

follows from the definition of $d$. Finally,

$$d(y_{i,j,K}, y_{i',j',K'}) + d(y_{i',j',K'}, y_{i'',j'',K''})$$
$$\geq d(y_{i,j,K}, y_{i'',j'',K''})$$

holds since

$$\max(d(x_{j,K}, y_{i,j,K}), d(x_{j',K'}, y_{i',j',K'}))$$
$$+ \max(d(x_{j',K'}, y_{i',j',K'}), d(x_{j'',K''}, y_{i'',j'',K''}))$$
$$\geq \max(d(x_{j,K}, y_{i,j,K}), d(x_{j'',K''}, y_{i'',j'',K''})) .$$

This proves that $d$ is a pseudo distance function.

Since we constructed $M$, we will check that the expected distance $ed_{P,d}$ satisfies $ed_{P,d}(\bigcup_{i \in I} \varphi_i^M) = e_I$. The distance between $x_{j,K}$ and $\bigcup_{i \in I} \varphi_i^M$ is as follows.

$$\begin{aligned}
d(x_{j,K}, \bigcup_{i \in I} \varphi_i^M) &= \min_{i \in I} \min_{j',K'} d(x_{j,K}, y_{i,j',K'}) \\
&= \begin{cases} 1 & \text{if } I \subset K \\ 0 & \text{else} . \end{cases}
\end{aligned}$$

where $j \notin K$ by definition. Therefore,

$$\begin{aligned}
&ed(\bigcup_{i \in I} \varphi_i^M) \\
&= \sum_{\omega \in \Omega} d(\omega, \bigcup_{i \in I} \varphi_i^M) \cdot P(\omega) \\
&= \sum_{j \in K^c} \Big[ \sum_{K \subset \{1,\cdots,n\}} d(x_{j,K}, \bigcup_{i \in I} \varphi_i^M) \cdot P(x_{j,K}) \Big] \\
&= \sum_{j \in K^c} \Big[ \sum_{K \subset \{1,\cdots,n\}, I \subset K} P(x_{j,K}) \Big] \\
&= \sum_{K \subset \{1,\cdots,n\}, I \subset K} \Big[ \sum_{j \in K^c} \frac{m_{K^c}}{|K^c|} \Big] \\
&= \sum_{K \subset \{1,\cdots,n\}, I \subset K} m_{K^c} \\
&= \sum_{K^c \subset I^c} m_{K^c} = \sum_{J \subset I^c} m_J = e_I .
\end{aligned}$$

$\square$

## 7 Discussion

We constructed reasoning systems on metric spaces using expected distance functions. These systems could be built on a metric space equipped with any kind of belief theoretic measures. The axioms we adopted for these systems were sound and complete. But symbols like "Pr" or "D" that represent probability or distance are not included in $\mathcal{L}^{ED}$. We showed that probability logic and the logic of expected distance are not more expressive than each other (Lee 2006). Furthermore, extended languages including some of those symbols are strictly more expressive than $\mathcal{L}^{ED}$. It is an interesting question whether there exist sets of complete axioms for those extended systems.

Reasoning with expected distance has many potential applications. One interesting application is a reasoning system for second order uncertainties (Gaifman 1986). Since uncertainty degrees are usually represented in metric spaces, we can adopt expected distance functions to represent second order uncertainties. As an example, an expression such as $ED("Prob(\varphi) = 0.5") = 0.1$ is an efficient representations when we represent a probability distribution over probabilities of $\varphi$. Since $Prob("Prob(\varphi) = 0.5")$ would always become zero, a purely probabilistic second order reasoning system should chose an interval such as $Prob(Prob(\varphi) \in [0.4, 0.6]) = 0.8$. The reasoning for this choice is not clear, and tends to lose more information in the process. There is no explicit reason why we should choose the previous expression instead of $Prob(Prob(\varphi) \in [0.2, 0.9]) = 0.9$. The representation with expected distance is more natural and intuitive.

**Acknowledgement** I thank Lawrence Moss for his support and guidance for this paper.